\documentclass[10pt,twocolumn,letterpaper]{article}

\usepackage[pagenumbers]{iccv} 
\usepackage{tabularx}  
\usepackage{amsmath}
\usepackage{multicol}
\usepackage[table,xcdraw]{xcolor}
\usepackage{kotex}
\usepackage{multirow}
\usepackage[linesnumbered,ruled,vlined]{algorithm2e} 
\usepackage{algpseudocode} 
\usepackage{makecell}

\definecolor{iccvblue}{rgb}{0.21,0.49,0.74}
\usepackage[pagebackref,breaklinks,colorlinks,allcolors=iccvblue]{hyperref}

\def\paperID{8257} 
\def\confName{ICCV}
\def\confYear{2025}

\title{InsideOut: Integrated RGB-Radiative Gaussian Splatting\\ for Comprehensive 3D Object Representation}

\author{
Jungmin Lee$^{1}$,\hspace{1cm}Seonghyuk Hong$^{3}$,\\ Juyong Lee$^{2}$,\hspace{1cm} Jaeyoon Lee$^{2}$,\hspace{1cm} Jongwon Choi$^{1,2}$\thanks{Corresponding author} \\
{\small $^1$Dept. of Advanced Imaging, GSAIM, Chung-Ang University, Republic of Korea}\\
{\small $^2$GS. of AI, Chung-Ang University, Republic of Korea}\\
{\small $^3$Cultural Heritage Conservation Science Center, National Research Institute of Cultural Heritage, Republic of Korea}\\
{\tt\small jngmnlee@vilab.cau.ac.kr, h123kr@korea.kr,}\\
{\tt\small\{jylee, 
leejaeyoon\}@vilab.cau.ac.kr,
\tt\small choijw@cau.ac.kr}
}

\begin{document}
\maketitle

\begin{abstract}
We introduce InsideOut, an extension of 3D Gaussian splatting (3DGS) that bridges the gap between high-fidelity RGB surface details and subsurface X-ray structures. The fusion of RGB and X-ray imaging is invaluable in fields such as medical diagnostics, cultural heritage restoration, and manufacturing. We collect new paired RGB and X-ray data, perform hierarchical fitting to align RGB and X-ray radiative Gaussian splats, and propose an X-ray reference loss to ensure consistent internal structures. InsideOut effectively addresses the challenges posed by disparate data representations between the two modalities and limited paired datasets. This approach significantly extends the applicability of 3DGS, enhancing visualization, simulation, and non-destructive testing capabilities across various domains.
\end{abstract}

\section{Introduction}
\label{sec:intro}
\begin{figure}[t!]
    \centering
    \includegraphics[width=\columnwidth]{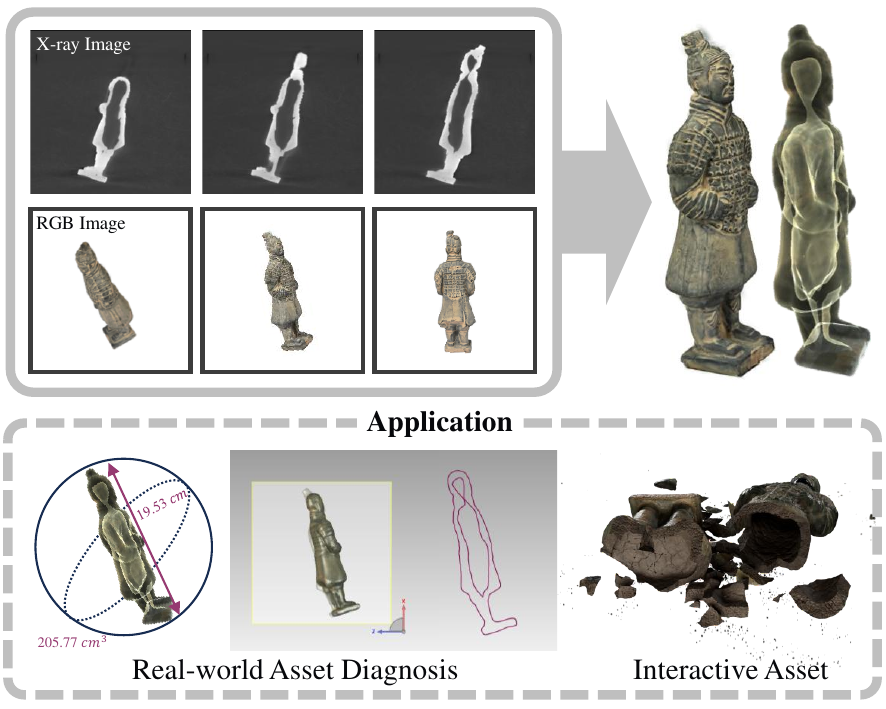}
    \caption{\textbf{Applications of InsideOut.} Our approach visualizes external appearance and internal structures, enabling analysis and physical simulation.}
    \label{fig:teaser}
    \vspace{-5mm}
\end{figure}


RGB and X-ray imaging offer strong complementary advantages in various fields, such as medical imaging~\cite{teoh2024advancing}, cultural heritage preservation~\cite{hess2015fusion}, and manufacturing~\cite{wang2023multimodal}. RGB images provide a high-fidelity visual appearance on external surfaces, but lack depth in internal structures. In contrast, X-ray imaging reveals subsurface structure, but fails to capture surface texture. The fusion of these modalities enables a more comprehensive understanding of both external appearance and internal structure, facilitating applications such as interactive simulations and non-destructive testing.

Despite their potential, the modality gap between RGB and X-ray imaging remains a critical challenge for effective fusion. This challenge primarily arises from the fundamental differences in how each modality captures visual information. X-rays penetrate objects and generate projection-based tomography with multi-layered structures, producing low-contrast grayscale images without external appearance details. In contrast, RGB imaging captures only the external appearance by reflecting light from object surfaces. These differences complicate feature alignment since X-ray images lack spatial correspondences with RGB features.

To address this challenge, we adopt 3D Gaussian splatting (3DGS)~\cite{kerbl20233d} as a novel approach for multi-modal data fusion. Optimizing both modalities within a unified 3D model allows for efficient spatial and visual alignment. The Gaussian splats in 3DGS encode position, opacity, and color parameters, enabling depth-aware modeling of multi-layered structures visible in X-ray images alongside RGB surface details. As a result, as shown in Fig.~\ref{fig:teaser}, the unified 3D model can enhance analysis and visualization capabilities in various applications, such as real-world asset diagnosis and interactive assets.

However, an integrated 3DGS model that simultaneously utilizes both RGB and X-ray data presents several challenges. First, existing datasets lack both RGB image sets and corresponding X-ray images, necessitating new data collection procedures. Second, the geometric representations of RGB and X-ray data differ significantly, making geometric alignment difficult to achieve within a single 3D model. Third, the two modalities capture fundamentally different information, with RGB data providing detailed texture and color information, whereas X-ray data only captures structural information.

To address these challenges, we propose InsideOut, which is an extension of 3DGS that jointly represents external appearance from RGB images and internal structures from X-ray images. For evaluation, we first collect a new dataset containing paired RGB images and X-ray images. We then perform a hierarchical fitting procedure to align the scale and position of initial RGB and radiative Gaussian splats, ensuring geometric consistency. Guided by a novel X-ray reference loss, cross-sectional X-ray images direct the training of both RGB and radiative Gaussian splats, enabling them to be accurately aligned by internal structures. Finally, as a post-processing step, we selectively duplicate fine-scale RGB splats into the radiative splats, transferring color information from nearby RGB splats. Consequently, the radiative Gaussian splats form a unified 3D model that preserves both detailed surface features and internal structures.
Our contributions are as follows:

\begin{itemize}
\item We propose a novel framework that integrates RGB and X-ray images into 3DGS, producing a unified 3D model with both external appearance and internal structures.

\item We develop a hierarchical fitting approach that bridges the geometric domain gap between RGB and X-ray 3D models, ensuring consistent position and scale alignment.

\item We introduce a new X-ray reference loss that reduces structural differences between RGB and X-ray 3D models based on cross-sectional X-ray guidance.

\item We construct and release a new dataset of paired RGB and X-ray images and demonstrate the effectiveness of our framework with the new dataset.
\end{itemize}

\begin{figure*}[t]
    \centering
    \includegraphics[width=\textwidth]{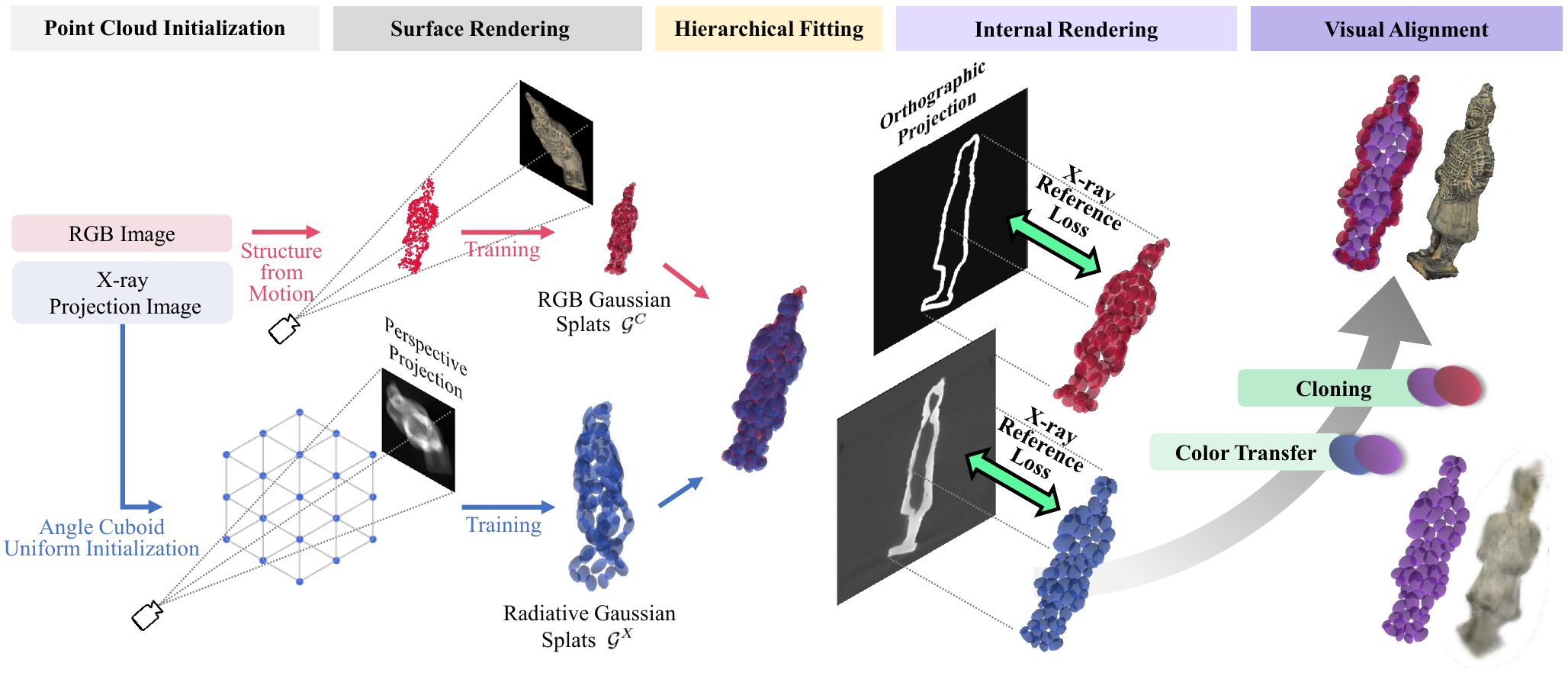}
    \caption{\textbf{Framework of InsideOut.} Our framework employs a five-stage pipeline to integrate RGB and X-ray images into a unified 3D representation. Starting with point cloud initialization from both modalities, the pipeline performs surface rendering to learn basic structures, hierarchical fitting to achieve geometric alignment, internal rendering with cross-sectional guidance for structural refinement, and visual alignment to transfer colors and details, ultimately producing a comprehensive 3D model capturing both external appearance and internal structures.}
    \label{fig:framework}
    \vspace{-5mm}
\end{figure*}

\section{Related works}
\label{sec:related works}
\subsection{Multi-modal Data Fusion}
\noindent\textbf{RGB Imaging-Based Fusion.}
RGB imaging preserves the texture of objects and finds wide utilization in various studies due to the natural similarity that RGB imaging shares with the human visual system~\cite{salau2021review, brenner2023rgb, kaur2021image}. Prior work has focused on visual interpretability by fusing RGB images with depth maps~\cite{marsh2022critical, wang2022rgb, wunsch2024data}, thermal images~\cite{alexander2022fusion, shopovska2019deep}, infrared images~\cite{sun2020infrared, liu2024rgb, luo2023infrared}, and multispectral images~\cite{sara2021hyperspectral, zhao2022deep}. For instance, combining RGB images with depth information enhances the realism of 3D reconstruction models~\cite{li2022high}, while recent work proposes implicit neural fusion of far-infrared images for improved 3D reconstruction~\cite{li2024implicit}. However, RGB imaging captures only surface information and fails to provide insight into internal structures.

\noindent\textbf{X-ray Imaging-Based Fusion.}
X-ray imaging can penetrate high-density materials, making it an essential modality for internal structure analysis~\cite{kozioziemski2023x}. Previous studies frequently combine X-ray imaging with other modalities such as MRI~\cite{yaqub2022deep, zhou2024edge, xu2024simultaneous}, structured light~\cite{pasko2020combining}, and time-of-flight (ToF) cameras~\cite{ehsani2019registration}. These methods integrate X-ray imaging with 3D structured light to improve the visualization of bone deformations~\cite{pasko2020combining} and with ToF cameras to enable cost-effective and accurate patient positioning~\cite{ehsani2019registration}. However, these studies primarily focus on improving visual representation rather than generating a complete 3D model, since X-ray imaging produces density-based projections from absorption patterns, posing inherent challenges in matching with surface-capturing modalities.

While X-ray and RGB imaging reveal internal and external structures respectively, the integration challenges stem from fundamental physical differences. Previous studies have applied multi-resolution wavelet transformation to fuse these modalities~\cite{nair2018multi} and calibrated geometry in X-ray images using RGB cameras~\cite{albiol2016geometrical}. Existing approaches focus on 2D image alignment, limited by geometric inconsistencies, camera parameter variations, and data interpretation differences. Full 3D reconstruction remains an unresolved problem, causing separate processing and independent analysis within the respective domains, even for the same object.

\subsection{Gaussian Splatting}
\noindent\textbf{RGB Imaging-Based 3DGS.}
Previous 3DGS methods have achieved remarkable success in RGB image rendering tasks, particularly in 3D object generation~\cite{qian2023magic123, yi2023gaussiandreamer, yang2024gaussianobject} and surface reconstruction~\cite{guedon2024sugar, huang20242d, yu2024gaussian, wolf2024gsmesh}. In 3D object generation, Gaussian representations enable advanced object creation from sparse inputs and effective object-background separation. However, these approaches focus on surface rendering without integrating internal structures, filling with Gaussian splat noise.
Recent work in surface reconstruction has evolved beyond 3D Gaussian object generation by incorporating depth information and normal vectors to enhance surface rendering quality. 
Although these studies have successfully converted Gaussian representations into external meshes, the methods still fail to address the limitations of internal structure representation.

\noindent\textbf{X-ray Imaging-Based 3DGS.}
Several studies have extended 3DGS to X-ray imaging, primarily for 3D sparse-view reconstruction to reduce radiation exposure~\cite{nikolakakis2024gaspct, cai2025radiative, gao2024ddgs}. While these approaches enable X-ray scene rendering, the methods limit 3DGS to internal structure representation. Similarly, volumetric methods using Gaussian kernels~\cite{li2023sparse, r2_gaussian} focus on density reconstruction but remain confined to X-ray imaging. Existing work has developed 3DGS separately within RGB and X-ray domains. In contrast, our study bridges this gap by integrating both modalities to represent internal and external object structures.

\section{Preliminaries}
\label{sec:preliminaries}

\subsection{RGB Gaussian Splatting}
The standard 3DGS~\cite{kerbl20233d} represents objects using a set of 3D Gaussian splats $\mathcal{G}^{C}$, where each individual Gaussian splat $G_i^C$ encapsulates the essential geometric and appearance characteristics:
\begin{equation}
G_i^C = \left\{ \mathbf{\mu}_i^C, \Sigma_i^C, \alpha_i^C, \mathbf{c}_i^C \right\}
\end{equation}
where \(\mathbf{\mu}_i^C \in \mathbb{R}^3\) and \(\Sigma_i^C \in \mathbb{R}^{3\times3}\) represent the center position and covariance matrix, respectively. \(\alpha_i^C \in [0,1]\) denotes the opacity, and \(\mathbf{c}_i^C \in \mathbb{R}^3\) represents the RGB color of the \(i\)-th Gaussian splat. The complete representation is expressed as:
\begin{equation}
\mathcal{G}^{C} = \left\{ G_i^C \mid i = 1, 2, \dots, N^C \right\},
\end{equation}
where $N^C$ denotes the total number of Gaussian splats in the representation.

We first generate a 3D point cloud using Structure from Motion (SfM)~\cite{schonberger2016structure} and define Gaussians centered at each point. During rendering, each Gaussian \(G_i^C\) is projected into 2D image space via the camera projection matrix \(\mathbf{P}\). The projected 2D Gaussian is characterized by its mean position \(\mathbf{p}_i = \mathbf{P}\mu_i\) and covariance matrix \(\mathbf{J}_i \Sigma_i \mathbf{J}_i^T\), where \(\mathbf{J}_i\) is the Jacobian of the camera projection at \(\mu_i^C\). The opacity \(\alpha_i^C\) determines the transparency of the projected Gaussian in the final rendered image.

The final 2D image is generated by differentiable rasterization, where all projected Gaussian splats contribute to the final rendering. The photometric loss between the projected and target images is minimized using gradient-based optimization, refining the position, covariance, opacity, and color of each Gaussian. 

\subsection{Radiative Gaussian Splatting}
We utilize X-Gaussian~\cite{cai2025radiative} to represent radiative Gaussian splats. Each individual radiative Gaussian splat $G_j^X$ is formally defined as:
\begin{equation}
G_j^X = \{\mu_j^X, \Sigma_j^X, \alpha_j^X, \mathbf{f}_j^X\},
\end{equation}
where $\mu_j^X$, $\Sigma_j^X$, and $\alpha_j^X$ denote the position, covariance, and opacity parameters, respectively, maintaining the same formulation as in standard RGB Gaussians. However, since X-ray imaging lacks color information, we use the feature vector $\mathbf{f}_j^X$, which represents the inherent radiative properties of the $j$-th Gaussian~\cite{cai2025radiative}.
The complete set of radiative Gaussian splats is expressed as:
\begin{equation}
\mathcal{G}^{X} = \{G_j^X \mid j=1,2,\ldots,N^X\},
\end{equation}
where $N^X$ denotes the total number of radiative Gaussian splats in the representation.

We use an angle cuboid uniform initialization (ACUI) to calculate intrinsic and extrinsic matrices from X-ray scanner parameters. ACUI uniformly samples 3D points within a cuboid that encompasses the scanned object to initialize radiative Gaussian point cloud center positions. The initialized 3D point cloud undergoes differentiable radiative rasterization to generate rendered images from given view directions. This process maintains isotropic properties by ensuring consistent radiation intensity across viewing angles, producing the projected image. We optimize radiative Gaussian splats using photometric loss by minimizing differences between rendered and X-ray projection images.

\section{Methodology}
\label{sec:method}
Fig.~\ref{fig:framework} illustrates the framework of InsideOut. Our framework integrates RGB and radiative Gaussian splats to create a comprehensive 3D object. First, we train RGB Gaussian splats using standard 3DGS~\cite{kerbl20233d} and radiative Gaussian splats using X-Gaussian~\cite{cai2025radiative} to learn the coarse internal and external structure. Second, we perform hierarchical fitting to align position, scale, and rotation between different modalities. Third, we use cross-sectional X-ray images to align surfaces and refine internal layers, clarifying boundaries and strengthening geometry. After geometric alignment, we achieve visual alignment by transferring colors to the X-ray 3D model and adding fine surface details from the RGB 3D model.

\subsection{Hierarchical Fitting}
\begin{figure}[t]
  \centering
  \includegraphics[width=\linewidth]{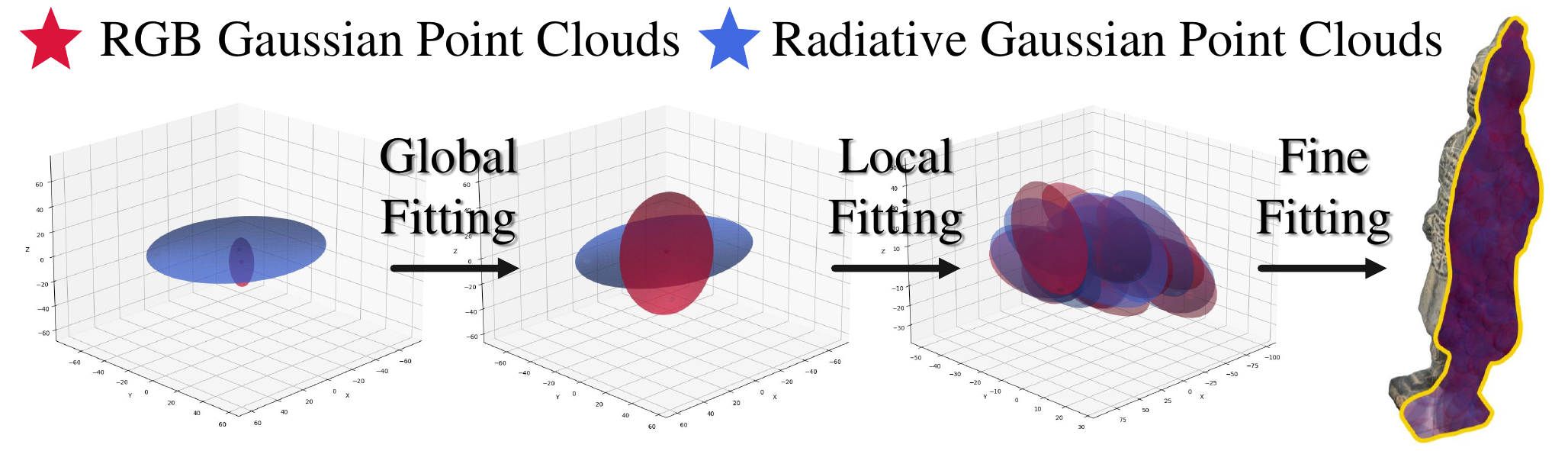}
      \caption{\textbf{Overview of hierarchical fitting.} The hierarchical fitting aligns RGB and radiative Gaussian point clouds through three progressive stages: global fitting for overall alignment, local fitting with clustering at multiple resolutions, and fine fitting for individual Gaussian alignment.}
  \label{fig:fitting}
  \vspace{-5mm}
\end{figure}
Based on the coarse Gaussian splats, we perform hierarchical fitting, as shown in Fig.~\ref{fig:fitting}. This process ensures consistent spatial correspondence between the RGB Gaussian splat set \( \mathcal{G}^{C} \) and the radiative Gaussian splat set \( \mathcal{G}^{X} \). By considering the mean and covariance, hierarchical fitting achieves effective alignment even when shape differences exist between the two modalities. The hierarchical fitting process consists of three progressive stages, namely global fitting, local fitting, and fine fitting. The overall algorithm flow is given in Alg.~\ref{alg:hier_fitting}.

In the global fitting stage, we approximate each Gaussian splat set, $\mathcal{G}^{C}$ and $\mathcal{G}^{X}$, as a single Gaussian $G^{C}$ and $G^{X}$. We compute the mean position and covariance for each modality $s \in \{C, X\}$ by averaging across all Gaussian splats within the set:
\begin{equation} 
    \bar{\mu}^{s} = \frac{1}{N_s} \sum_{l=1}^{N_s} \mu_l^{s}, \quad 
    \bar{\Sigma}^{s} = \frac{1}{N_s} \sum_{l=1}^{N_s} \Sigma_l^{s},
\end{equation}
where $N_s$ denotes the number of Gaussian splats in the set $\mathcal{G}^{s}$, and $\mu_l^{s}$ and $\Sigma_l^{s}$ represent the mean and covariance of the $l$-th Gaussian splat in the set, respectively. 

After computing the global mean and covariance, we perform affine alignment to transform $G^C$ to match $G^X$. SfM-generated RGB point clouds operate in relative coordinate systems, while X-ray point clouds use absolute physical coordinates from scanner parameters. To bridge this coordinate gap, we compute the affine transformation by aligning centroids, normalizing scales, and aligning orientations through PCA-derived rotation. The composite transformation is then applied to all individual Gaussian splats, updating their positions and covariances accordingly.

\begin{algorithm}[t]
\caption{Hierarchical Fitting}
\label{alg:hier_fitting}
\KwIn{$\mathcal{G}^{X}=\{G_j\}$, $\mathcal{G}^{C}=\{G_i\}$ \\ 
\quad \quad (each has $N_X$ and $N_C$ Gaussians)}
\KwOut{Aligned $\mathcal{G}^C$}
\nl Compute $(\bar{\mu}^X,\bar{\Sigma}^X) \leftarrow \text{GlobalMean}(\mathcal{G}^X)$ and 
    $(\bar{\mu}^C,\bar{\Sigma}^C) \leftarrow \text{GlobalMean}(\mathcal{G}^C)$\;
\nl Construct global Gaussians $G^X(\mathbf{x}),\,G^C(\mathbf{x})$ using $(\bar{\mu}^X, \bar{\Sigma}^X)$ and $(\bar{\mu}^C, \bar{\Sigma}^C)$\;
\nl Align $G^C(\mathbf{x})$ to $G^X(\mathbf{x})$ using ICP~\cite{besl1992method}\;
\nl \For{$K$ in $\{N/10,\; N/2,\; N\}$}{
\nl \quad Cluster $\mathcal{G}^X$ and $\mathcal{G}^C$ into $K$ clusters: \\
    \quad $C^X \leftarrow \text{KMeansCluster}(\mathcal{G}^X, K)$, \\ 
    \quad $C^C \leftarrow \text{KMeansCluster}(\mathcal{G}^C, K)$\;
\nl \quad \For{each cluster $C_k$}{
\nl \quad \quad Compute cluster-level mean and covariance: \\
\quad \quad $(\bar{\mu}^X_{k}, \bar{\Sigma}^X_{k}) \leftarrow \text{Mean}(C^X_k)$, \\
\quad \quad $(\bar{\mu}^C_k, \bar{\Sigma}^C_k) \leftarrow \text{Mean}(C^C_k)$\;
\nl \quad \quad Align $C^C_k$ to $C^X_k$ using ICP\;
    }
}
\nl Perform fine alignment on all Gaussians in $\mathcal{G}^C$\;
\end{algorithm}

After global fitting, the alignment is progressively refined through hierarchical clustering. The Gaussian splats are first grouped into $\lfloor N/10 \rfloor$ clusters, then into $\lfloor N/2 \rfloor$ clusters, before finalizing alignment using all $N$ Gaussian splats. We apply K-means clustering~\cite{hartigan1979algorithm} to each modality separately, grouping spatially neighboring Gaussian splats. Each Gaussian splat $G_l \in \mathcal{G}^s$ is assigned to its nearest cluster center, forming clusters $C^s_k$:
\begin{equation}
C^s_k = \left\{ G_l^s \;\Bigm|\;\| \mu_l^s - \bar{\mu}_{k}^s \|_2 = \min_{k'\in K} \| \mu_l^s - \bar{\mu}_{k'}^s \|_2 \right\},
\end{equation}
where $C^s_k$ denotes the $k$-th cluster, $\mu^s_l$ represents the mean position of Gaussian splat $G_l$, $\bar{\mu}_{k'}$ denotes the centroid of cluster $k'$, and $K$ represents all possible cluster indices. 

After clustering, the process refines the alignment by approximating each cluster using its mean and covariance as:
\begin{equation}
    \tilde{\mu}^s_k = \frac{1}{|C^s_k|} \sum_{G_{l} \in C^s_k} \mu^s_{l}, \quad 
    \tilde{\Sigma}^s_k = \frac{1}{|C^s_k|} \sum_{G_{l} \in C^s_k} \Sigma^s_{l},
\end{equation}
where $|C^s_k|$ denotes the number of Gaussian splats in cluster $C^s_k$. This strategy aligns both modalities spatially while preserving their geometric structures.
\subsection{Internal Rendering}
To sharpen the multi-layer boundaries of 3D models, we introduce cross-sectional images obtained through orthographic projection of X-ray volumetric data as pseudo-ground truth (GT). These images provide fine-grained supervision for internal alignment, reinforcing consistency between RGB Gaussian splats and radiative Gaussian splats throughout training. The pseudo-GT is generated by slicing the X-ray volume at 15mm intervals along the axial and coronal planes, refining the spatial distribution of radiative Gaussian splats.

For RGB Gaussian splats, we apply a Canny edge detector~\cite{canny1986computational} to extract surface contours from the sliced images. When multiple edge layers are detected, we identify the outermost surface boundary by computing bounding boxes for each edge contour and selecting the one with the largest area. The extracted surface edges serve as pseudo-GT for training RGB Gaussian splats. This approach guides the splats to align along the object's surface while keeping the internal region empty. During optimization, we update only the geometric parameters $\mu_i^C$, $\Sigma_i^C$, and $\alpha_i^C$ while excluding the color parameter $\mathbf{c}_i^C$, since the pseudo-GT provides only grayscale edge information.

To optimize the internal structure, we introduce an X-ray reference loss that combines multiple loss terms to achieve accurate layer separation and structural consistency. The total X-ray reference loss is:
\begin{equation}    
    \mathcal{L} = (1 - \lambda_{\text{s}}) \cdot \mathcal{L}_{\text{L1}} +  \lambda_{\text{s}} \cdot \mathcal{L}_{\text{D-SSIM}} + \lambda_{\text{z}} \cdot \mathcal{L}_{\text{zero-one}},
\end{equation}
where $\mathcal{L}_{\text{L1}}$ and $\mathcal{L}_{\text{D-SSIM}}$ are photometric losses~\cite{wang2004image} that compare the rendered image with the X-ray image. The parameter $\lambda_{\text{s}}$ controls the balance between L1 and D-SSIM losses for photometric reconstruction quality. Additionally, $\mathcal{L}_{\text{zero-one}}$~\cite{xu2022point, lombardi2019neural, guedon2024sugar, jiang2024gaussianshader} constrains the opacity values $\alpha_i$ of Gaussian splats to converge to either 0 or 1 during training, with weight $\lambda_{\text{z}}$. This constraint enforces the formation of thin planar structures at specific layer locations, sharpening layer boundaries and accurately capturing radiation intensity. As a result, this approach removes internal noise from the RGB 3D model while promoting well-defined layer structures in the X-ray 3D model.

\subsection{Visual Alignment}
Once the RGB and X-ray 3D models are fully aligned, we transfer the surface color from the RGB Gaussian splats to the radiative Gaussian splats. Then, we duplicate the fine splats from the RGB object to the X-ray 3D model to create a richly detailed surface.

\noindent\textbf{Color Transfer.}
To assign color to radiative Gaussian splats, we apply k-Nearest Neighbor (k-NN) matching by finding the closest RGB Gaussian splat for each radiative Gaussian splat:
\begin{equation} 
    \mathbf{c}_j^X = \mathbf{c}^C_{\operatorname*{argmin}_{i \in \{1,2,\ldots,N_C\}} \|\mu_j^X - \mu_i^C\|_2},
\end{equation}
\noindent
where the color attributes are transferred from the RGB Gaussian splat with minimum distance to each radiative Gaussian splat. Since radiative Gaussian splats are initially defined without color attributes, this process assigns color parameters $\mathbf{c}^X_j$ to ensure they reflect the surface appearance from RGB Gaussian splats, preserving the external visual appearance of the object.

\noindent\textbf{Detail Cloning.}
To enhance surface detail while preserving structural consistency, we selectively duplicate high-frequency splats from the RGB 3D model into the X-ray 3D model. Gaussian splats are classified based on their covariance values, where smaller covariance indicates more detailed structures. We retain the bottom 95\% of RGB Gaussian splats ranked by covariance magnitude, discarding the top 5\% with large covariance values. First, we extract the largest eigenvalue from each RGB Gaussian splat's covariance matrix:
\begin{equation}
\Lambda^C = \{\lambda_\text{max}(\Sigma_i^C) \mid i=1,2,\dots,N^C\},
\end{equation}
where $\lambda_{\max}(\Sigma_i^C)$ denotes the largest eigenvalue of the covariance matrix $\Sigma_i^C$ for the $i$-th RGB Gaussian splat. The threshold is computed as the $p$-th percentile of the eigenvalue distribution:
\begin{equation}
T_p = \text{Quantile}(\Lambda^C, p/100),
\end{equation}
where $p$ represents the percentile value. The detailed Gaussian splats are selected based on the threshold:
\begin{equation}
{G}^{\text{detail}} = \{G_i^{C} \mid \lambda_\text{max}(\Sigma_i^C)\leq T_p, i=1,2,\dots,N^C\},
\end{equation}
Empirical results show that setting $p = 95$ works well. These detailed Gaussian splats are incorporated into the radiative Gaussian splat set:
\begin{equation} 
    \hat{\mathcal{G}}^{X} = \mathcal{G}^{X} \cup \mathcal{G}^{\text{detail}}.
\end{equation}
where $\hat{\mathcal{G}}^{X}$ denotes the enhanced radiative Gaussian splat set with detailed surface features. This integration enhances surface fidelity without disrupting the internal structure.

\section{Experiments}
\label{sec:experiments}

\begin{figure}[t]
    \vspace{-0.5cm}
    \centering    \includegraphics[width=\columnwidth]{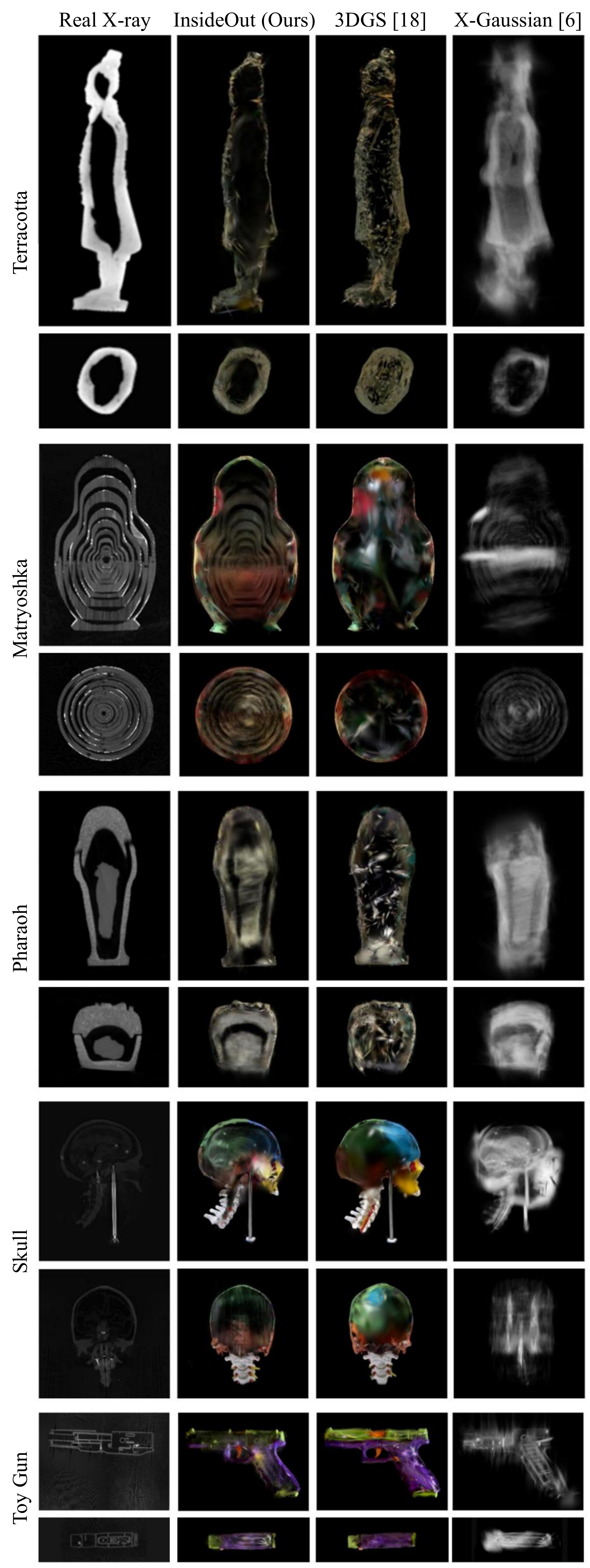}
    \caption{\textbf{Qualitative comparison of cross-sectional 3D models.} Coronal slices (upper) and axial slices (lower) are listed.}
    \label{fig:total_rendering}
    \vspace{-1.5cm}
\end{figure}


We implement InsideOut using PyTorch~\cite{paszke2019pytorch} with CUDA~\cite{sanders2010cuda}. The training pipeline consists of two optimization stages: surface appearance and internal structure optimization. For surface rendering, standard 3DGS~\cite{kerbl20233d} was trained for 20k iterations, while X-Gaussian~\cite{cai2025radiative} was trained for 5k iterations. The resulting Gaussian point cloud was then used as initial point cloud for internal rendering, which was further optimized for 20k iterations. We apply loss weights $\lambda_{\text{s}} = 0.2$ and $\lambda_{\text{z}}=0.005$. We employ Adam optimizer~\cite{kingma2014adam} with an initial learning rate of 5e-4. All experiments were conducted on an NVIDIA A6000 GPU.

\subsection{Dataset.}
We constructed a novel RGB-X-ray paired dataset encompassing five objects across diverse application domains including medical imaging, cultural heritage preservation, and manufacturing quality control. Our dataset includes Matryoshka, Pharaoh, Terracotta, Skull, and Toy Gun, as detailed in Tab.~\ref{tab:dataset}. For RGB data collection, we captured 90-644 images per object using smartphones and DSLR cameras in natural lighting conditions with over 60\% overlap between consecutive views to ensure comprehensive coverage. Image resolutions range from 3024×1848 to 5712×4284 pixels depending on the capture device. For X-ray data acquisition, we used a cone-beam CT scanner with 360° rotation. Acquisition parameters were optimized per object based on material properties, with voltage ranging from 185-330 kV and current from 0.49-1.50 mA, and 2048 detector elements with 2016-2048 Z-slices.

\begin{table}[t]
\centering
\resizebox{\columnwidth}{!}{%
\begin{tabular}{@{}l|c|c|c|c|c@{}}
\specialrule{.1em}{.05em}{.05em}
&\textbf{Matryoshka} & \textbf{Pharaoh} & \textbf{Terracotta} & \textbf{Skull} & \textbf{Toy Gun}\\ 
\specialrule{.1em}{.05em}{.05em}
\# RGB Images & 555 & 644 & 210 & 90 & 416\\  
Resolution (pixels) & $4000 \times 1848$ & $4000 \times 1848$ & $4000 \times 4032$ & $3024 \times 1848$ & $5712 \times 4284$\\  
\specialrule{.1em}{.05em}{.05em}
\# X-ray Projections & 360 & 360 & 2160 & 360&360 \\  
\# Detector Elements & 2048 & 2048 & 2048 & 2048& 2048\\  
\# Z Slices & 2048 & 2048 & 2016 & 2048& 2048\\  
Voltage (kV) & 230.00 & 330.00 & 185.00 & 230.00& 230.00 \\  
Current (mA) & 1.35 & 1.50 & 0.49 & 1.35& 1.35\\  
\specialrule{.1em}{.05em}{.05em}
\end{tabular}}
\caption{\textbf{Dataset specifications.} RGB image collection details (top) and X-ray acquisition parameters (bottom) for five objects.}
\label{tab:dataset} 
\vspace{-5mm}
\end{table}
\begin{table*}[t!]
\centering
\resizebox{\linewidth}{!}{%
\begin{tabular}{@{}l|ccc|ccc|ccc|ccc|ccc@{}}
\specialrule{.1em}{.05em}{.05em} 
 & \multicolumn{3}{c|}{\textbf{Matryoshka}} 
 & \multicolumn{3}{c|}{\textbf{Pharaoh}}
 & \multicolumn{3}{c|}{\textbf{Terracotta}}
 & \multicolumn{3}{c|}{\textbf{Skull}}
 & \multicolumn{3}{c}{\textbf{Toy Gun}} \\[1pt]
 & PIQE $\downarrow$ & PSNR $\uparrow$ & SSIM $\uparrow$ 
 & PIQE $\downarrow$ & PSNR $\uparrow$ & SSIM $\uparrow$ 
 & PIQE $\downarrow$ & PSNR $\uparrow$ & SSIM $\uparrow$
 & PIQE $\downarrow$ & PSNR $\uparrow$ & SSIM $\uparrow$
 & PIQE $\downarrow$ & PSNR $\uparrow$ & SSIM $\uparrow$\\
\specialrule{.1em}{.05em}{.05em}
3DGS~\cite{kerbl20233d}      
  & 54.63 & 16.25 & 0.60  
  & 43.71 & 13.66 & 0.62 
  & 30.35 & 13.47 & 0.46
  & 51.71  &16.24  &0.61   
  & 44.63  &18.48  &0.61 \\

3DGS-MCMC~\cite{kheradmand20243d}     
  & 50.84 & 16.12 & 0.59 
  & 41.63 & 13.82 & 0.64
  & 30.36 & 13.98 & 0.46 
  & 50.35 & 16.81 & 0.61  
  & 43.22 & 18.68 & 0.61 \\

DoF-Gaussian~\cite{shen2025dof}
  &53.62 & 16.59 & 0.60
  &43.25 & 13.85 & 0.63
  &31.83 & 14.31 & 0.47
  &50.79 & 16.71 & 0.62
  &44.29 & 18.36 & 0.61 \\
  
MaskGaussian~\cite{liu2025maskgaussian}   
  & 54.34& 16.43 & 0.60
  & 42.15& 13.41 & 0.63
  & 30.02 & 13.82& 0.46
  &56.66 & 16.35 & 0.61 
  &43.58 & 18.26 & 0.61 \\

Analytic-Splatting~\cite{liang2024analytic}
  & 52.71 & 16.67 &0.60
  & 41.29 & 13.37 &0.62
  & 31.85 & 13.24 &0.47
  & 50.39 & 15.87 & 0.61 
  & 43.82 & 17.58 & 0.60 \\

RS-NeRF~\cite{niu2024rs}
  & 52.96 & 16.88 & 0.60
  & 44.08 & 13.64 & 0.62
  & 32.75 & 13.57 & 0.47
  & 52.32 & 15.41 & 0.61
  & 44.67 & 17.59 & 0.61 \\

NAF~\cite{zha2022naf}
  &59.63&17.03&0.61
  &64.39&14.21&0.64
  &61.51&14.88&0.49
  &67.82 &18.64 &0.63
  &60.32 &18.23 &0.63  \\
  
X-Gaussian~\cite{cai2025radiative}    
  & 29.82 & 17.46 & \textbf{0.61}  
  & 33.33 & 14.78 & \textbf{0.64}
  & \textbf{25.11} & 15.20 & 0.49 
  &23.71  & 19.42 & 0.64
  &21.27 & 19.53 & 0.64  \\

\rowcolor[HTML]{ECF4FF}
InsideOut (Ours)    
  & \textbf{25.28} & \textbf{18.86} & \textbf{0.61}  
  & \textbf{30.35} & \textbf{15.14} & \textbf{0.64} 
  & 25.35 & \textbf{19.56} & \textbf{0.55}  
  & \textbf{20.01} & \textbf{20.33} & \textbf{0.65} 
  & \textbf{20.74} & \textbf{20.87} & \textbf{0.66}  \\
\specialrule{.1em}{.05em}{.05em}
\end{tabular}%
}
\caption{\textbf{Comparison of quality for internal detail.} We present PIQE, PSNR, and SSIM scores for internal detail quality across five objects. All methods are evaluated on cross-sectional views for internal structure assessment. The best results are in bold, and our method is shown in blue.}
\label{tab:internal_detail}
\vspace{-5mm}
\end{table*}

\begin{figure*}[t]
    \centering
    \includegraphics[width=\textwidth]{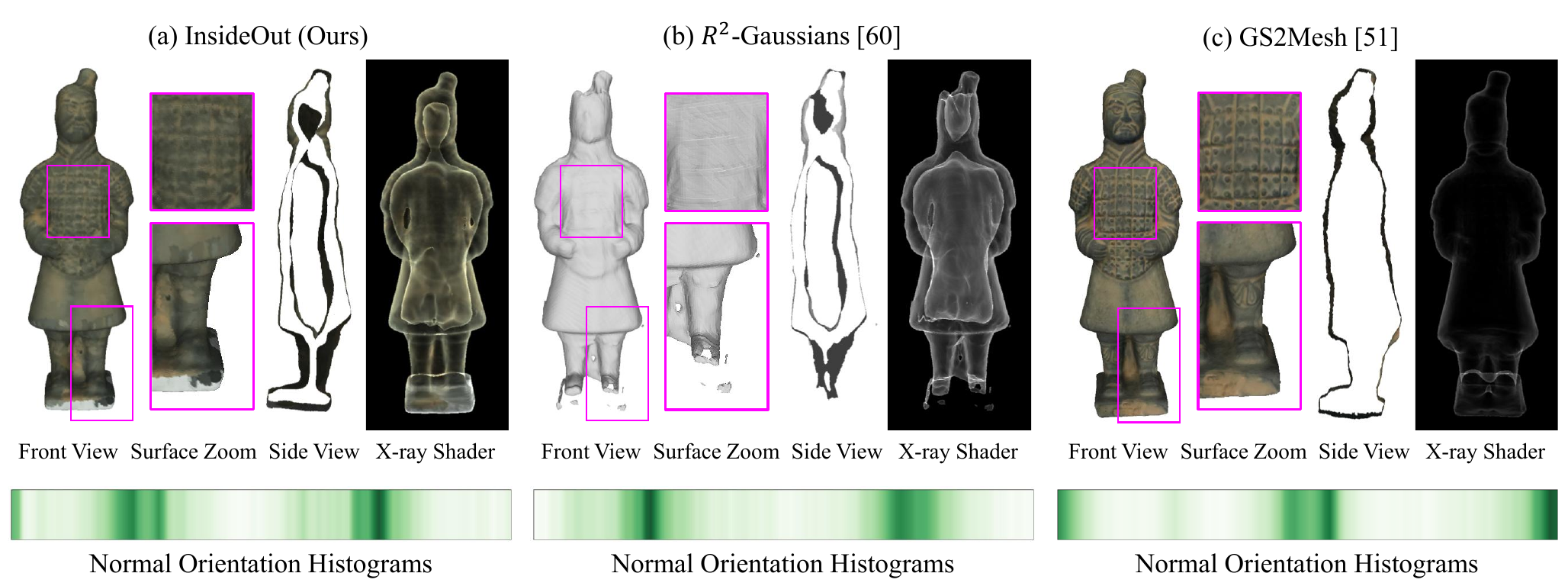}
    \caption{\textbf{Comparison of surface and internal structure in mesh reconstruction.} Mesh reconstruction quality is compared across multiple views: front view, surface zoom, side view, and X-ray shader rendering. Pink boxes indicate zoomed regions. Normal orientation histograms below each method illustrate the distribution of surface normals.}
    \label{fig:mesh_normal}
    \vspace{-5mm}
\end{figure*}
\begin{figure}[t]
    \centering
    \includegraphics[width=\columnwidth]{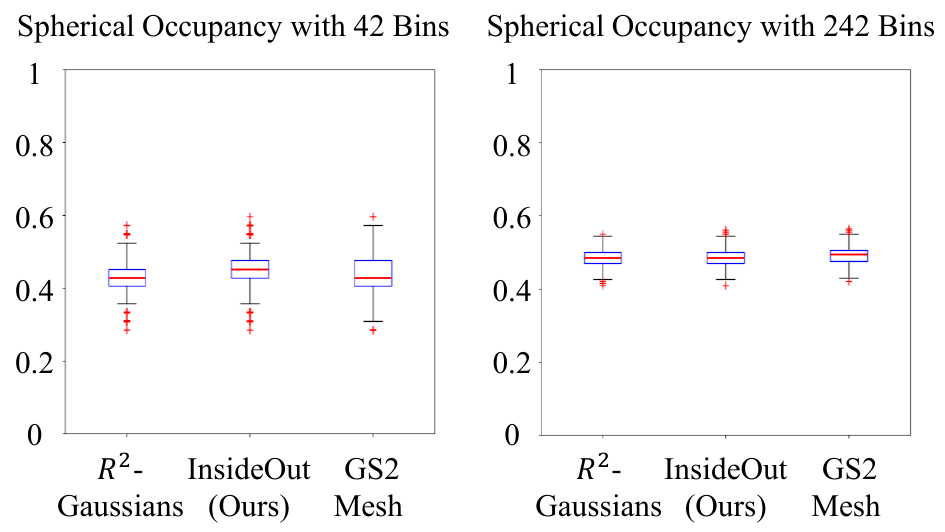}
    \caption{\textbf{Box plots for distribution of spherical occupancy.} We compare two different binning approaches: 42 bins and 242 bins.}
    \label{fig:box_plot}
    \vspace{-5mm}
\end{figure}

\subsection{Internal Rendering}
Fig.~\ref{fig:total_rendering} presents a qualitative comparison of internal rendering, highlighting structural clarity and noise reduction improvements. Unlike 3DGS, which produces internal noise and scattered Gaussian splats, InsideOut eliminated these artifacts and concentrated the splats along the object's surface, creating a cleaner representation. Additionally, InsideOut enhanced the visibility of internal layers that X-Gaussian blurred, producing sharper and more well-defined structural contours. The results demonstrate that InsideOut achieved superior internal detail preservation while mitigating noise, thereby improving overall rendering fidelity.

According to Tab.~\ref{tab:internal_detail}, our method achieved the lowest perceptual image quality evaluator (PIQE)~\cite{venkatanath2015blind} score across all datasets, demonstrating superior internal detail preservation. Since no GT images exist for simultaneously evaluating both the internal and external structures of a 3D model, we employ a no-reference image quality assessment. We slice the output into cross-sections and use PIQE metrics to evaluate the perceptual quality.

Tab.~\ref{tab:internal_detail} presents the quantitative evaluation of InsideOut in cross-sectional views, where we compute PSNR and SSIM against reference X-ray images. On average, InsideOut improved PSNR by 1.67 dB over X-Gaussian and by 3.33 dB over 3DGS, demonstrating significant enhancement in reconstruction quality. Similarly, SSIM increased by 0.018 compared to X-Gaussian and by 0.042 compared to 3DGS, indicating improved structural consistency and perceptual quality.

\subsection{Mesh Reconstruction}
We construct 3D mesh models to assess the quality and applicability of InsideOut. We generate meshes from R\textsuperscript{2}-Gaussian~\cite{r2_gaussian}, a radiative volume reconstruction method, by applying Marching Cubes~\cite{lorensen1998marching}. We then compare the results with GS2Mesh~\cite{wolf2024gsmesh}, an RGB-based mesh reconstruction approach. For InsideOut, we apply the density voxelizer from R\textsuperscript{2}-Gaussian and then generate meshes using Marching Cubes.

Fig.~\ref{fig:mesh_normal} qualitatively evaluates the three meshes, offering insights into structural fidelity. InsideOut captured finer geometric details than radiative meshes while preserving the overall shape of the object. Additionally, rendering with MeshLab's X-ray Shader~\cite{LocalChapterEvents:ItalChap:ItalianChapConf2008:129-136} enabled comparison with RGB-based meshes, demonstrating that InsideOut reconstructed both the external surface and internal structure with high fidelity. As shown in Fig.~\ref{fig:teaser}, our method supports real-world asset diagnosis and physical simulation, enabling cross-sectional analysis and quantitative assessment of internal structures. However, since we employed a voxelization method for radiative Gaussian splats, the surface texture tends to be of lower quality compared to GS2Mesh.

To quantify fine-grained surface variations, we introduce the normal orientation histogram (NOH) and spherical occupancy~\cite{lamb2023fantastic}. Spherical occupancy measures the distribution of surface normals by partitioning the unit sphere into discrete regions $n$. We assign each normal to the nearest reference point, forming the NOH, which captures the directional distribution. We then compute spherical occupancy as the fraction of occupied regions in the NOH.

Fig.~\ref{fig:mesh_normal} presents the NOH for three mesh models, illustrating the distribution of surface normals across directional space. R\textsuperscript{2}-Gaussian exhibits minimal variation in normal directions, indicating a simple structure, whereas InsideOut shows a more diverse distribution of normals across different directions. The histograms provide insight into the geometric complexity of each 3D model by revealing variations in normal orientation density.

Fig.~\ref{fig:box_plot} presents spherical occupancy values for R\textsuperscript{2}-Gaussian, InsideOut, and GS2Mesh at bin resolutions of 42 and 242. At 42 bins, InsideOut achieved the highest occupancy (0.443 mean, 0.452 median), followed by GS2Mesh (0.437 mean, 0.429 median) and R²-Gaussian (0.431 mean, 0.429 median). At 242 bins, GS2Mesh led with 0.476 mean and 0.475 median, while InsideOut recorded 0.471 for both metrics, and R²-Gaussian remained lowest at 0.468 mean and 0.466 median. These findings indicate that InsideOut captures more diverse normal distributions than R²-Gaussian and represents internal layer normals that GS2Mesh cannot capture. 


\begin{figure}[t]
    \centering
    \includegraphics[width=\columnwidth]{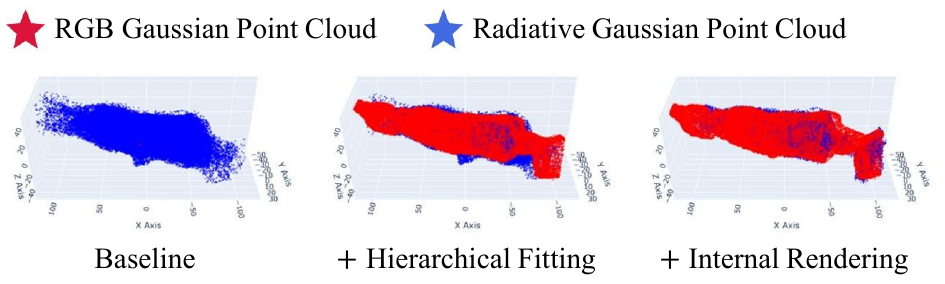}
    \caption{\textbf{Qualitative results of ablation study.} The figure shows the progressive improvement achieved by each module: baseline (no module), hierarchical alignment only, and hierarchical alignment with internal rendering.}
    \label{fig:ablation}
    \vspace{-5mm}
\end{figure}
\subsection{Ablation Study}
\noindent\textbf{Geometrical Alignment.}
Fig.~\ref{fig:ablation} shows that hierarchical fitting improved initial correspondence, while internal rendering refined the structure. The leftmost visualization shows the initial state where RGB Gaussian point clouds (red) appear much smaller than radiative Gaussian point clouds (blue) due to scale differences, making red points barely visible. With hierarchical fitting (middle), the two point clouds achieved substantial alignment. The rightmost shows the complete method, achieving optimal alignment. The progression demonstrates that hierarchical fitting provides primary alignment benefits, while internal rendering further refines the geometry.


\begin{table}[t!]
\centering
\resizebox{\linewidth}{!}{%
\begin{tabular}{@{}cc|c|c|c|c|c@{}}
\specialrule{.1em}{.05em}{.05em}
\multicolumn{2}{c|}{\textbf{Module}} 
& \multicolumn{1}{c|}{\multirow{3}{*}{\textbf{Matryoshka}}} 
& \multicolumn{1}{c|}{\multirow{3}{*}{\textbf{Pharaoh}}} 
& \multicolumn{1}{c|}{\multirow{3}{*}{\textbf{Terracotta}}}
& \multicolumn{1}{c|}{\multirow{3}{*}{\textbf{Skull}}}
& \multicolumn{1}{c}{\multirow{3}{*}{\textbf{Toy Gun}}}\\ 
\makecell{Hierarchical \\ Alignment} & \makecell{Internal \\ Rendering} &  &  &  &  &  \\ 
\specialrule{.1em}{.05em}{.05em} 
           &  & 38.56 & 27.93 & 21.29 & 722.45 & 23.02 \\
\checkmark &  & 8.41 & 12.52 & 11.44 & 9.53 & 4.59 \\
\checkmark & \checkmark & \textbf{6.45} & \textbf{4.10} & \textbf{2.42} & \textbf{7.95} & \textbf{3.42} \\
\specialrule{.1em}{.05em}{.05em} 
\end{tabular}}
\caption{\textbf{Comparison of chamfer distance for geometrical alignment.} The table presents progressive improvement achieved by each module: baseline (no module), hierarchical alignment only, and hierarchical alignment with internal rendering.}
\label{tab:alignment}
\vspace{-5mm}
\end{table}

Tab.~\ref{tab:alignment} compares chamfer distance (CD) to evaluate the effects of hierarchical fitting and internal rendering. The results show that hierarchical fitting significantly reduced CD values across all 3D models, substantially enhancing initial alignment. Internal rendering further refined geometric accuracy, leading to the lowest CD values. To compute CD, we extracted surface points from both point clouds using Hidden Point Removal (HPR)~\cite{katz2007direct}. This approach ensured that only visible surface points contributed to the alignment evaluation while excluding internal points from radiative Gaussian point clouds.

\section{Conclusion}
\label{sec:conclusion}
We introduce InsideOut, a novel 3DGS framework that integrates RGB and X-ray data to simultaneously capture an object's external surface and internal structure. By leveraging 3DGS, our approach effectively aligns these two modalities within a unified 3D representation. To achieve this, we first apply hierarchical fitting to structure Gaussian splats based on their position and scale, ensuring geometric consistency. Next, we refine the internal representation using cross-sectional X-ray guidance, allowing the model to delineate internal boundaries with higher precision. Finally, we transfer color and surface details from RGB Gaussians to radiative Gaussians, producing a comprehensive 3D model that preserves both external textures and internal structures. This advancement supports applications such as diagnostics, virtual simulations, and industrial inspection, making InsideOut a powerful tool for real-world use.

\vspace{3mm}
\footnotesize 
\noindent\textbf{Acknowledgements:}
This research was partly supported by Culture, Sports and Tourism R\&D Program through the Korea Creative Content Agency grant funded by Ministry of Culture, Sports and  Tourism in 2024 (Project Name : Developing Professionals for R\&D in Contents Production Based on Generative AI and Cloud, Project Number: RS-2024-00352578, Contribution Rate: 60\%), Culture Technology R\&D Program in 2023 (Project Name: Development of Intelligent Heritage Platform for Leading of Standardization on Digital Cultural Heritage, Project Number: RS-2023-00219579), and the Institute of Information \& Communications Technology Planning \& Evaluation (IITP) grant funded by the Korea government (MSIT) [RS-2021-II211341, Artificial Intelligence Graduate School Program (Chung-Ang University)].
\normalsize

{
    \small
    \bibliographystyle{ieeenat_fullname}
    \bibliography{main}

\begin{thebibliography}{61}
\providecommand{\natexlab}[1]{#1}
\providecommand{\url}[1]{\texttt{#1}}
\expandafter\ifx\csname urlstyle\endcsname\relax
  \providecommand{\doi}[1]{doi: #1}\else
  \providecommand{\doi}{doi: \begingroup \urlstyle{rm}\Url}\fi

\bibitem[Albiol et~al.(2016)Albiol, Corbi, and Albiol]{albiol2016geometrical}
Francisco Albiol, Alberto Corbi, and Alberto Albiol.
\newblock Geometrical calibration of x-ray imaging with rgb cameras for 3d reconstruction.
\newblock \emph{IEEE transactions on medical imaging}, 35\penalty0 (8):\penalty0 1952--1961, 2016.

\bibitem[Alexander et~al.(2022)Alexander, Hoskere, Narazaki, Maxwell, and Spencer~Jr]{alexander2022fusion}
Quincy~G Alexander, Vedhus Hoskere, Yasutaka Narazaki, Andrew Maxwell, and Billie~F Spencer~Jr.
\newblock Fusion of thermal and rgb images for automated deep learning based crack detection in civil infrastructure.
\newblock \emph{AI in Civil Engineering}, 1\penalty0 (1):\penalty0 3, 2022.

\bibitem[Besl and McKay(1992)]{besl1992method}
P.~J. Besl and N.~D. McKay.
\newblock A method for registration of 3-d shapes.
\newblock \emph{IEEE Transactions on Pattern Analysis and Machine Intelligence}, 14\penalty0 (2):\penalty0 239--256, 1992.

\bibitem[Brenner et~al.(2023)Brenner, Reyes, Susnjak, and Barczak]{brenner2023rgb}
Martin Brenner, Napoleon~H Reyes, Teo Susnjak, and Andre~LC Barczak.
\newblock Rgb-d and thermal sensor fusion: a systematic literature review.
\newblock \emph{IEEE Access}, 2023.

\bibitem[Cai et~al.(2025)Cai, Liang, Wang, Wang, Zhang, Yang, Zhou, and Yuille]{cai2025radiative}
Yuanhao Cai, Yixun Liang, Jiahao Wang, Angtian Wang, Yulun Zhang, Xiaokang Yang, Zongwei Zhou, and Alan Yuille.
\newblock Radiative gaussian splatting for efficient x-ray novel view synthesis.
\newblock In \emph{European Conference on Computer Vision}, pages 283--299. Springer, 2025.

\bibitem[Canny(1986)]{canny1986computational}
John Canny.
\newblock A computational approach to edge detection.
\newblock \emph{IEEE Transactions on pattern analysis and machine intelligence}, \penalty0 (6):\penalty0 679--698, 1986.

\bibitem[Cignoni et~al.(2008)Cignoni, Callieri, Corsini, Dellepiane, Ganovelli, and Ranzuglia]{LocalChapterEvents:ItalChap:ItalianChapConf2008:129-136}
Paolo Cignoni, Marco Callieri, Massimiliano Corsini, Matteo Dellepiane, Fabio Ganovelli, and Guido Ranzuglia.
\newblock {MeshLab: an Open-Source Mesh Processing Tool}.
\newblock In \emph{Eurographics Italian Chapter Conference}. The Eurographics Association, 2008.

\bibitem[Ehsani et~al.(2019)Ehsani, Pouladian, Toosizadeh, and Aledavood]{ehsani2019registration}
Omid Ehsani, M Pouladian, S Toosizadeh, and A Aledavood.
\newblock Registration and fusion of 3d surface data from ct and tof camera for position verification in radiotherapy.
\newblock \emph{SN Applied Sciences}, 1\penalty0 (11):\penalty0 1347, 2019.

\bibitem[Gao et~al.(2024)Gao, Planche, Zheng, Chen, Chen, and Wu]{gao2024ddgs}
Zhongpai Gao, Benjamin Planche, Meng Zheng, Xiao Chen, Terrence Chen, and Ziyan Wu.
\newblock Ddgs-ct: Direction-disentangled gaussian splatting for realistic volume rendering.
\newblock \emph{Advances in Neural Information Processing Systems}, 37:\penalty0 39281--39302, 2024.

\bibitem[Gu{\'e}don and Lepetit(2024)]{guedon2024sugar}
Antoine Gu{\'e}don and Vincent Lepetit.
\newblock Sugar: Surface-aligned gaussian splatting for efficient 3d mesh reconstruction and high-quality mesh rendering.
\newblock In \emph{Proceedings of the IEEE/CVF Conference on Computer Vision and Pattern Recognition}, pages 5354--5363, 2024.

\bibitem[Hartigan and Wong(1979)]{hartigan1979algorithm}
John~A Hartigan and Manchek~A Wong.
\newblock Algorithm as 136: A k-means clustering algorithm.
\newblock \emph{Journal of the royal statistical society. series c (applied statistics)}, 28\penalty0 (1):\penalty0 100--108, 1979.

\bibitem[Hess et~al.(2015)Hess, Petrovic, Meyer, Rissolo, and Kuester]{hess2015fusion}
Michael Hess, Vid Petrovic, Dominique Meyer, Dominique Rissolo, and Falko Kuester.
\newblock Fusion of multimodal three-dimensional data for comprehensive digital documentation of cultural heritage sites.
\newblock In \emph{2015 Digital Heritage}, pages 595--602. IEEE, 2015.

\bibitem[Huang et~al.(2024)Huang, Yu, Chen, Geiger, and Gao]{huang20242d}
Binbin Huang, Zehao Yu, Anpei Chen, Andreas Geiger, and Shenghua Gao.
\newblock 2d gaussian splatting for geometrically accurate radiance fields.
\newblock In \emph{ACM SIGGRAPH 2024 Conference Papers}, pages 1--11, 2024.

\bibitem[Jiang et~al.(2024)Jiang, Tu, Liu, Gao, Long, Wang, and Ma]{jiang2024gaussianshader}
Yingwenqi Jiang, Jiadong Tu, Yuan Liu, Xifeng Gao, Xiaoxiao Long, Wenping Wang, and Yuexin Ma.
\newblock Gaussianshader: 3d gaussian splatting with shading functions for reflective surfaces.
\newblock In \emph{Proceedings of the IEEE/CVF Conference on Computer Vision and Pattern Recognition}, pages 5322--5332, 2024.

\bibitem[Katz et~al.(2007)Katz, Tal, and Basri]{katz2007direct}
Sagi Katz, Ayellet Tal, and Ronen Basri.
\newblock Direct visibility of point sets.
\newblock In \emph{ACM SIGGRAPH 2007 papers}, pages 24--es. 2007.

\bibitem[Kaur et~al.(2021)Kaur, Koundal, and Kadyan]{kaur2021image}
Harpreet Kaur, Deepika Koundal, and Virender Kadyan.
\newblock Image fusion techniques: a survey.
\newblock \emph{Archives of computational methods in Engineering}, 28\penalty0 (7):\penalty0 4425--4447, 2021.

\bibitem[Kerbl et~al.(2023)Kerbl, Kopanas, Leimk{\"u}hler, and Drettakis]{kerbl20233d}
Bernhard Kerbl, Georgios Kopanas, Thomas Leimk{\"u}hler, and George Drettakis.
\newblock 3d gaussian splatting for real-time radiance field rendering.
\newblock \emph{ACM Transactions on Graphics}, 42\penalty0 (4):\penalty0 139--1, 2023.

\bibitem[Kheradmand et~al.(2024)Kheradmand, Rebain, Sharma, Sun, Tseng, Isack, Kar, Tagliasacchi, and Yi]{kheradmand20243d}
Shakiba Kheradmand, Daniel Rebain, Gopal Sharma, Weiwei Sun, Yang-Che Tseng, Hossam Isack, Abhishek Kar, Andrea Tagliasacchi, and Kwang~Moo Yi.
\newblock 3d gaussian splatting as markov chain monte carlo.
\newblock \emph{Advances in Neural Information Processing Systems}, 37:\penalty0 80965--80986, 2024.

\bibitem[Kingma(2014)]{kingma2014adam}
Diederik~P Kingma.
\newblock Adam: A method for stochastic optimization.
\newblock \emph{arXiv preprint arXiv:1412.6980}, 2014.

\bibitem[Kozioziemski et~al.(2023)Kozioziemski, Bachmann, Do, and Tommasini]{kozioziemski2023x}
B Kozioziemski, B Bachmann, A Do, and R Tommasini.
\newblock X-ray imaging methods for high-energy density physics applications.
\newblock \emph{Review of Scientific Instruments}, 94\penalty0 (4), 2023.

\bibitem[Lamb et~al.(2023)Lamb, Palmer, Molloy, Banerjee, and Banerjee]{lamb2023fantastic}
Nikolas Lamb, Cameron Palmer, Benjamin Molloy, Sean Banerjee, and Natasha~Kholgade Banerjee.
\newblock Fantastic breaks: A dataset of paired 3d scans of real-world broken objects and their complete counterparts.
\newblock In \emph{Proceedings of the IEEE/CVF Conference on Computer Vision and Pattern Recognition}, pages 4681--4691, 2023.

\bibitem[Li et~al.(2022)Li, Gao, Wu, Liu, and Shen]{li2022high}
Jianwei Li, Wei Gao, Yihong Wu, Yangdong Liu, and Yanfei Shen.
\newblock High-quality indoor scene 3d reconstruction with rgb-d cameras: A brief review.
\newblock \emph{Computational Visual Media}, 8\penalty0 (3):\penalty0 369--393, 2022.

\bibitem[Li et~al.(2024)Li, Xie, Sakurada, Sagawa, and Oishi]{li2024implicit}
Xiangjie Li, Shuxiang Xie, Ken Sakurada, Ryusuke Sagawa, and Takeshi Oishi.
\newblock Implicit neural fusion of rgb and far-infrared 3d imagery for invisible scenes.
\newblock In \emph{2024 IEEE/RSJ International Conference on Intelligent Robots and Systems}, pages 12501--12508. IEEE, 2024.

\bibitem[Li et~al.(2023)Li, Fu, Zhao, Jin, and Zhou]{li2023sparse}
Yingtai Li, Xueming Fu, Shang Zhao, Ruiyang Jin, and S~Kevin Zhou.
\newblock Sparse-view ct reconstruction with 3d gaussian volumetric representation.
\newblock \emph{arXiv preprint arXiv:2312.15676}, 2023.

\bibitem[Liang et~al.(2024)Liang, Zhang, Hu, Zhu, Feng, and Jia]{liang2024analytic}
Zhihao Liang, Qi Zhang, Wenbo Hu, Lei Zhu, Ying Feng, and Kui Jia.
\newblock Analytic-splatting: Anti-aliased 3d gaussian splatting via analytic integration.
\newblock In \emph{European conference on computer vision}, pages 281--297. Springer, 2024.

\bibitem[Liu et~al.(2024)Liu, Li, Lu, and Huang]{liu2024rgb}
Ke Liu, Hang Li, Dongdong Lu, and Huabing Huang.
\newblock Rgb-infrared image fusion and classification based on tensor decomposition.
\newblock In \emph{Conference on Spectral Technology and Applications}, pages 928--936. SPIE, 2024.

\bibitem[Liu et~al.(2025)Liu, Zhong, Zhan, Xu, and Sun]{liu2025maskgaussian}
Yifei Liu, Zhihang Zhong, Yifan Zhan, Sheng Xu, and Xiao Sun.
\newblock Maskgaussian: Adaptive 3d gaussian representation from probabilistic masks.
\newblock In \emph{Proceedings of the Computer Vision and Pattern Recognition Conference}, pages 681--690, 2025.

\bibitem[Lombardi et~al.(2019)Lombardi, Simon, Saragih, Schwartz, Lehrmann, and Sheikh]{lombardi2019neural}
Stephen Lombardi, Tomas Simon, Jason Saragih, Gabriel Schwartz, Andreas Lehrmann, and Yaser Sheikh.
\newblock Neural volumes: Learning dynamic renderable volumes from images.
\newblock \emph{arXiv preprint arXiv:1906.07751}, 2019.

\bibitem[Lorensen and Cline(1998)]{lorensen1998marching}
William~E Lorensen and Harvey~E Cline.
\newblock Marching cubes: A high resolution 3d surface construction algorithm.
\newblock In \emph{Seminal graphics: pioneering efforts that shaped the field}, pages 347--353. 1998.

\bibitem[Luo and Luo(2023)]{luo2023infrared}
Yongyu Luo and Zhongqiang Luo.
\newblock Infrared and visible image fusion: Methods, datasets, applications, and prospects.
\newblock \emph{Applied Sciences}, 13\penalty0 (19):\penalty0 10891, 2023.

\bibitem[Marsh et~al.(2022)Marsh, Sadka, and Bahai]{marsh2022critical}
Benedict Marsh, Abdul~Hamid Sadka, and Hamid Bahai.
\newblock A critical review of deep learning-based multi-sensor fusion techniques.
\newblock \emph{Sensors}, 22\penalty0 (23):\penalty0 9364, 2022.

\bibitem[Nair and Singh(2018)]{nair2018multi}
Rekha~R Nair and Tripty Singh.
\newblock Multi-sensor, multi-modal medical image fusion for color images: A multi-resolution approach.
\newblock In \emph{2018 Tenth international conference on advanced computing (ICoAC)}, pages 249--254. IEEE, 2018.

\bibitem[Nikolakakis et~al.(2024)Nikolakakis, Gupta, Vengosh, Bui, and Marinescu]{nikolakakis2024gaspct}
Emmanouil Nikolakakis, Utkarsh Gupta, Jonathan Vengosh, Justin Bui, and Razvan Marinescu.
\newblock Gaspct: Gaussian splatting for novel ct projection view synthesis.
\newblock \emph{arXiv preprint arXiv:2404.03126}, 2024.

\bibitem[Niu et~al.(2024)Niu, Chen, Zhan, Li, Ji, and Zheng]{niu2024rs}
Muyao Niu, Tong Chen, Yifan Zhan, Zhuoxiao Li, Xiang Ji, and Yinqiang Zheng.
\newblock Rs-nerf: Neural radiance fields from rolling shutter images.
\newblock In \emph{European Conference on Computer Vision}, pages 163--180. Springer, 2024.

\bibitem[Pa{\'s}ko and Glinkowski(2020)]{pasko2020combining}
S{\l}awomir Pa{\'s}ko and Wojciech Glinkowski.
\newblock Combining 3d structured light imaging and spine x-ray data improves visualization of the spinous lines in the scoliotic spine.
\newblock \emph{Applied Sciences}, 11\penalty0 (1):\penalty0 301, 2020.

\bibitem[Paszke et~al.(2019)Paszke, Gross, Massa, Lerer, Bradbury, Chanan, Killeen, Lin, Gimelshein, Antiga, et~al.]{paszke2019pytorch}
Adam Paszke, Sam Gross, Francisco Massa, Adam Lerer, James Bradbury, Gregory Chanan, Trevor Killeen, Zeming Lin, Natalia Gimelshein, Luca Antiga, et~al.
\newblock Pytorch: An imperative style, high-performance deep learning library.
\newblock \emph{Advances in neural information processing systems}, 32, 2019.

\bibitem[Qian et~al.(2024)Qian, Mai, Hamdi, Ren, Siarohin, Li, Lee, Skorokhodov, Wonka, Tulyakov, and Ghanem]{qian2023magic123}
Guocheng Qian, Jinjie Mai, Abdullah Hamdi, Jian Ren, Aliaksandr Siarohin, Bing Li, Hsin-Ying Lee, Ivan Skorokhodov, Peter Wonka, Sergey Tulyakov, and Bernard Ghanem.
\newblock Magic123: One image to high-quality 3d object generation using both 2d and 3d diffusion priors.
\newblock In \emph{The Twelfth International Conference on Learning Representations}, 2024.

\bibitem[Salau et~al.(2021)Salau, Jain, and Eneh]{salau2021review}
Ayodeji~Olalekan Salau, Shruti Jain, and Joy~Nnenna Eneh.
\newblock A review of various image fusion types and transform.
\newblock \emph{Indonesian Journal of Electrical Engineering and Computer Science}, 24\penalty0 (3):\penalty0 1515--1522, 2021.

\bibitem[Sanders(2010)]{sanders2010cuda}
Jason Sanders.
\newblock \emph{CUDA by Example: An Introduction to General-Purpose GPU Programming}.
\newblock Addison-Wesley Professional, 2010.

\bibitem[Sara et~al.(2021)Sara, Mandava, Kumar, Duela, and Jude]{sara2021hyperspectral}
Dioline Sara, Ajay~Kumar Mandava, Arun Kumar, Shiny Duela, and Anitha Jude.
\newblock Hyperspectral and multispectral image fusion techniques for high resolution applications: A review.
\newblock \emph{Earth Science Informatics}, 14\penalty0 (4):\penalty0 1685--1705, 2021.

\bibitem[Schonberger and Frahm(2016)]{schonberger2016structure}
Johannes~L Schonberger and Jan-Michael Frahm.
\newblock Structure-from-motion revisited.
\newblock In \emph{Proceedings of the IEEE Conference on Computer Vision and Pattern Recognition}, pages 4104--4113, 2016.

\bibitem[Shen et~al.(2025)Shen, Liu, Sun, Li, Cao, Li, and Loy]{shen2025dof}
Liao Shen, Tianqi Liu, Huiqiang Sun, Jiaqi Li, Zhiguo Cao, Wei Li, and Chen~Change Loy.
\newblock Dof-gaussian: Controllable depth-of-field for 3d gaussian splatting.
\newblock In \emph{Proceedings of the Computer Vision and Pattern Recognition Conference}, pages 26462--26471, 2025.

\bibitem[Shopovska et~al.(2019)Shopovska, Jovanov, and Philips]{shopovska2019deep}
Ivana Shopovska, Ljubomir Jovanov, and Wilfried Philips.
\newblock Deep visible and thermal image fusion for enhanced pedestrian visibility.
\newblock \emph{Sensors}, 19\penalty0 (17):\penalty0 3727, 2019.

\bibitem[Sun et~al.(2020)Sun, Zhang, and Xiong]{sun2020infrared}
Changqi Sun, Cong Zhang, and Naixue Xiong.
\newblock Infrared and visible image fusion techniques based on deep learning: A review.
\newblock \emph{Electronics}, 9\penalty0 (12):\penalty0 2162, 2020.

\bibitem[Teoh et~al.(2024)Teoh, Dong, Zuo, Lai, Hasikin, and Wu]{teoh2024advancing}
Jing~Ru Teoh, Jian Dong, Xiaowei Zuo, Khin~Wee Lai, Khairunnisa Hasikin, and Xiang Wu.
\newblock Advancing healthcare through multimodal data fusion: a comprehensive review of techniques and applications.
\newblock \emph{PeerJ Computer Science}, 10:\penalty0 e2298, 2024.

\bibitem[Venkatanath et~al.(2015)Venkatanath, Praneeth, Bh, Channappayya, and Medasani]{venkatanath2015blind}
Narasimhan Venkatanath, D Praneeth, Maruthi~Chandrasekhar Bh, Sumohana~S Channappayya, and Swarup~S Medasani.
\newblock Blind image quality evaluation using perception based features.
\newblock In \emph{2015 twenty first national conference on communications}, pages 1--6. IEEE, 2015.

\bibitem[Wang et~al.(2022)Wang, Wang, Che, Xu, Qiao, Qi, Feng, and Tang]{wang2022rgb}
Haowen Wang, Mingyuan Wang, Zhengping Che, Zhiyuan Xu, Xiuquan Qiao, Mengshi Qi, Feifei Feng, and Jian Tang.
\newblock Rgb-depth fusion gan for indoor depth completion.
\newblock In \emph{Proceedings of the IEEE/CVF Conference on Computer Vision and Pattern Recognition}, pages 6209--6218, 2022.

\bibitem[Wang et~al.(2023)Wang, Peng, Zhang, Yi, Wang, and Wang]{wang2023multimodal}
Yue Wang, Jinlong Peng, Jiangning Zhang, Ran Yi, Yabiao Wang, and Chengjie Wang.
\newblock Multimodal industrial anomaly detection via hybrid fusion.
\newblock In \emph{Proceedings of the IEEE/CVF Conference on Computer Vision and Pattern Recognition}, pages 8032--8041, 2023.

\bibitem[Wang et~al.(2004)Wang, Bovik, Sheikh, and Simoncelli]{wang2004image}
Zhou Wang, Alan~C Bovik, Hamid~R Sheikh, and Eero~P Simoncelli.
\newblock Image quality assessment: from error visibility to structural similarity.
\newblock \emph{IEEE transactions on image processing}, 13\penalty0 (4):\penalty0 600--612, 2004.

\bibitem[Wolf et~al.(2024)Wolf, Bracha, and Kimmel]{wolf2024gsmesh}
Yaniv Wolf, Amit Bracha, and Ron Kimmel.
\newblock {GS}2{M}esh: Surface reconstruction from {G}aussian splatting via novel stereo views.
\newblock In \emph{European Conference on Computer Vision}, 2024.

\bibitem[Wunsch et~al.(2024)Wunsch, Tenorio, Anding, Golomoz, and Notni]{wunsch2024data}
Lennard Wunsch, Christian~G{\"o}rner Tenorio, Katharina Anding, Andrei Golomoz, and Gunther Notni.
\newblock Data fusion of rgb and depth data with image enhancement.
\newblock \emph{Journal of Imaging}, 10\penalty0 (3):\penalty0 73, 2024.

\bibitem[Xu et~al.(2022)Xu, Xu, Philip, Bi, Shu, Sunkavalli, and Neumann]{xu2022point}
Qiangeng Xu, Zexiang Xu, Julien Philip, Sai Bi, Zhixin Shu, Kalyan Sunkavalli, and Ulrich Neumann.
\newblock Point-nerf: Point-based neural radiance fields.
\newblock In \emph{Proceedings of the IEEE/CVF Conference on Computer Vision and Pattern Recognition}, pages 5438--5448, 2022.

\bibitem[Xu et~al.(2024)Xu, Li, Jie, and Tan]{xu2024simultaneous}
Yushen Xu, Xiaosong Li, Yuchan Jie, and Haishu Tan.
\newblock Simultaneous tri-modal medical image fusion and super-resolution using conditional diffusion model.
\newblock In \emph{International Conference on Medical Image Computing and Computer-Assisted Intervention}, pages 635--645. Springer, 2024.

\bibitem[Yang et~al.(2024)Yang, Li, Fang, Liang, Xie, Zhang, Shen, and Tian]{yang2024gaussianobject}
Chen Yang, Sikuang Li, Jiemin Fang, Ruofan Liang, Lingxi Xie, Xiaopeng Zhang, Wei Shen, and Qi Tian.
\newblock Gaussianobject: High-quality 3d object reconstruction from four views with gaussian splatting.
\newblock \emph{ACM Transactions on Graphics}, 2024.

\bibitem[Yaqub et~al.(2022)Yaqub, Jinchao, Arshid, Ahmed, Zhang, Nawaz, and Mahmood]{yaqub2022deep}
Muhammad Yaqub, Feng Jinchao, Kaleem Arshid, Shahzad Ahmed, Wenqian Zhang, Muhammad~Zubair Nawaz, and Tariq Mahmood.
\newblock Deep learning-based image reconstruction for different medical imaging modalities.
\newblock \emph{Computational and Mathematical Methods in Medicine}, 2022\penalty0 (1):\penalty0 8750648, 2022.

\bibitem[Yi et~al.(2024)Yi, Fang, Wang, Wu, Xie, Zhang, Liu, Tian, and Wang]{yi2023gaussiandreamer}
Taoran Yi, Jiemin Fang, Junjie Wang, Guanjun Wu, Lingxi Xie, Xiaopeng Zhang, Wenyu Liu, Qi Tian, and Xinggang Wang.
\newblock Gaussiandreamer: Fast generation from text to 3d gaussians by bridging 2d and 3d diffusion models.
\newblock In \emph{Proceedings of the IEEE/CVF Conference on Computer Vision and Pattern Recognition}, 2024.

\bibitem[Yu et~al.(2024)Yu, Sattler, and Geiger]{yu2024gaussian}
Zehao Yu, Torsten Sattler, and Andreas Geiger.
\newblock Gaussian opacity fields: Efficient adaptive surface reconstruction in unbounded scenes.
\newblock \emph{ACM Transactions on Graphics}, 2024.

\bibitem[Zha et~al.(2022)Zha, Zhang, and Li]{zha2022naf}
Ruyi Zha, Yanhao Zhang, and Hongdong Li.
\newblock Naf: neural attenuation fields for sparse-view cbct reconstruction.
\newblock In \emph{International Conference on Medical Image Computing and Computer-Assisted Intervention}, pages 442--452. Springer, 2022.

\bibitem[Zha et~al.(2024)Zha, Lin, Cai, Cao, Zhang, and Li]{r2_gaussian}
Ruyi Zha, Tao~Jun Lin, Yuanhao Cai, Jiwen Cao, Yanhao Zhang, and Hongdong Li.
\newblock R$^2$-gaussian: Rectifying radiative gaussian splatting for tomographic reconstruction.
\newblock In \emph{Advances in Neural Information Processing Systems}, 2024.

\bibitem[Zhao et~al.(2022)Zhao, Kumar, Banoth, Marathi, Rajalakshmi, Rewald, Ninomiya, and Guo]{zhao2022deep}
Jiangsan Zhao, Ajay Kumar, Balaji~Naik Banoth, Balram Marathi, Pachamuthu Rajalakshmi, Boris Rewald, Seishi Ninomiya, and Wei Guo.
\newblock Deep-learning-based multispectral image reconstruction from single natural color rgb image—enhancing uav-based phenotyping.
\newblock \emph{Remote Sensing}, 14\penalty0 (5):\penalty0 1272, 2022.

\bibitem[Zhou et~al.(2024)Zhou, Zhang, Xu, Wang, and Khalvati]{zhou2024edge}
Meng Zhou, Yuxuan Zhang, Xiaolan Xu, Jiayi Wang, and Farzad Khalvati.
\newblock Edge-enhanced dilated residual attention network for multimodal medical image fusion.
\newblock In \emph{2024 IEEE International Conference on Bioinformatics and Biomedicine (BIBM)}, pages 4108--4111. IEEE, 2024.

\end{thebibliography}
}

\end{document}